\documentclass[letterpaper, 10 pt, conference]{ieeeconf}

\usepackage{graphicx}
\usepackage{caption}
\usepackage{subcaption}
\usepackage{amsmath,amsfonts,amssymb}
\usepackage{array}
\usepackage{colortbl}
\usepackage{boldline}
\usepackage{algorithmic}
\usepackage[dvipsnames]{xcolor}
\usepackage{algorithm}
\usepackage{array}
\usepackage{textcomp}
\usepackage{stfloats}
\usepackage{url}
\usepackage{verbatim}

\usepackage[utf8]{inputenc} 
\usepackage[T1]{fontenc}
\usepackage{url}            

\usepackage{booktabs}
\usepackage{caption}

\usepackage{amsfonts}
\usepackage{nicefrac}       
\usepackage{microtype}      
\usepackage{xcolor}      
\usepackage{graphicx}

\usepackage{amsthm}

\usepackage{mathtools}
\usepackage{tabularx}
\usepackage{framed}
\usepackage{bbm}
\usepackage{bm}
\usepackage{xspace}
\usepackage{amssymb}
\usepackage{multirow}
\usepackage{physics}
\usepackage{siunitx}
\usepackage{algorithm}
\usepackage{algorithmic}

\usepackage{cite}

\usepackage{nccmath}

\usepackage{colortbl}

\usepackage{placeins}

\usepackage{subcaption}

\usepackage{lipsum}

\usepackage{thmtools}
\usepackage{thm-restate}

\makeatletter
\let\NAT@parse\undefined
\makeatother
\usepackage{hyperref}       \hypersetup{
    colorlinks,
    linkcolor={blue!90!black},
    citecolor={blue!90!black},
urlcolor={blue!95!black}
}

\usepackage[capitalise]{cleveref}       \usepackage{crossreftools}  \usepackage[all=normal,paragraphs=tight,floats=tight,mathspacing=tight,wordspacing=tight,tracking=tight]{savetrees}

\definecolor{Gray}{gray}{0.85}
\definecolor{LightCyan}{rgb}{0.88,1,1}
\definecolor{darkpurple}{RGB}{128,0,128}
\definecolor{darkbrown}{RGB}{51, 25, 0}
\definecolor{LightGray}{gray}{0.93}
\definecolor{mygreen}{rgb}{0.0, 0.5, 0.0}

\IEEEoverridecommandlockouts                              

\title{\LARGE \bf
Weathering Ongoing Uncertainty: Learning and Planning in a Time-Varying Partially Observable Environment
}

\author{Gokul Puthumanaillam, Xiangyu Liu, Negar Mehr and Melkior Ornik
\thanks{This work was supported by U.S. Army ERDC under cooperative agreement W9132T2220004, AFOSR grant FA9550-23-1-0131, ONR grant N00014-23-1-2505, NASA grant 80NSSC21K1030, and NSF CNS-2218759}
\thanks{This work was conducted while all authors were affiliated with the Department of Aerospace Engineering and the
Coordinated Science Laboratory, University of Illinois Urbana-Champaign,
Urbana, USA.
        {\tt\small \{gokulp2,mornik\}}@illinois.edu}
        \thanks{
        Subsequent to the completion of this work, Xiangyu Liu transitioned to the University of Cyprus {\tt\small \{liu.xiangyu\}}@ucy.ac.cy, and Negar Mehr to the University of California, Berkeley {\tt\small \{negar\}}@berkeley.edu}
}

\begin{document}

\maketitle
\thispagestyle{empty}
\pagestyle{empty}

\begin{abstract}

Optimal decision-making
presents a significant challenge for autonomous systems operating in uncertain, stochastic and time-varying environments. 
Environmental variability over time can significantly impact the system's optimal decision making strategy for mission completion.
To model such environments, our work combines the previous notion of Time-Varying Markov Decision Processes (TVMDP) with partial observability and introduces Time-Varying Partially Observable Markov Decision Processes (TV-POMDP). 
We propose a two-pronged approach to accurately estimate and plan within the TV-POMDP: 
1) Memory Prioritized State Estimation (MPSE), which leverages weighted memory to provide more accurate time-varying transition estimates; and 2) an MPSE-integrated planning strategy that optimizes long-term rewards while accounting for temporal constraint.
 We validate the proposed framework and algorithms using simulations and hardware, with robots exploring a partially observable, time-varying environments. Our results demonstrate superior performance over standard methods, highlighting the framework's effectiveness in stochastic, uncertain, time-varying domains.

\end{abstract}

\section{Introduction}
\begin{figure}[h]
    \centering
    \begin{subfigure}[b]{0.23\textwidth}
        \centering
        \includegraphics[width=\textwidth]{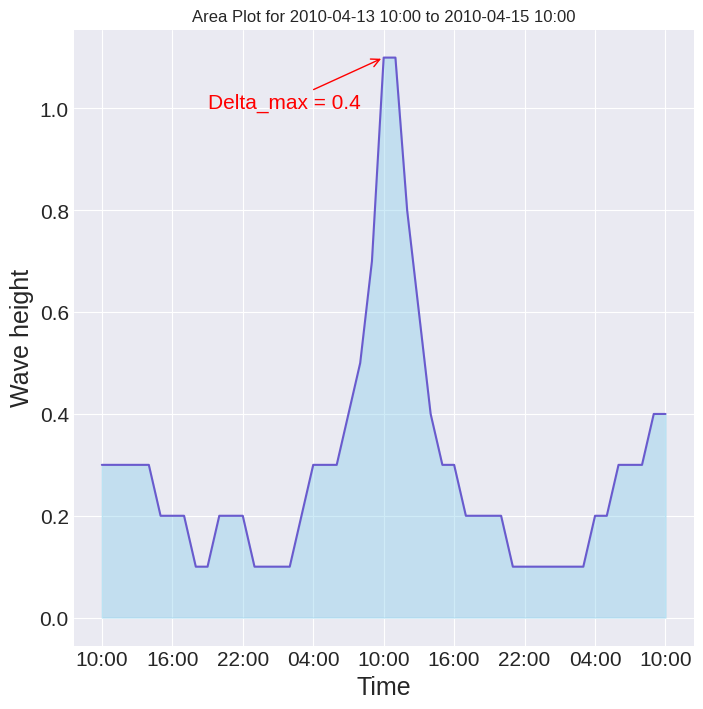}
        \caption{Area plot depicting the variation in wave heights (disturbance) over a 48-hour period. The data is part of a dataset that includes various environmental factors \cite{Huang2017}. The \textcolor{red}{red arrow} annotation indicates the maximal rate of change (\( \Delta_{\text{max}} \)) in wave height. }
        \label{fig:figure1}
    \end{subfigure}
    \hfill
    \begin{subfigure}[b]{0.24\textwidth}
        \centering
        \includegraphics[width=\textwidth]{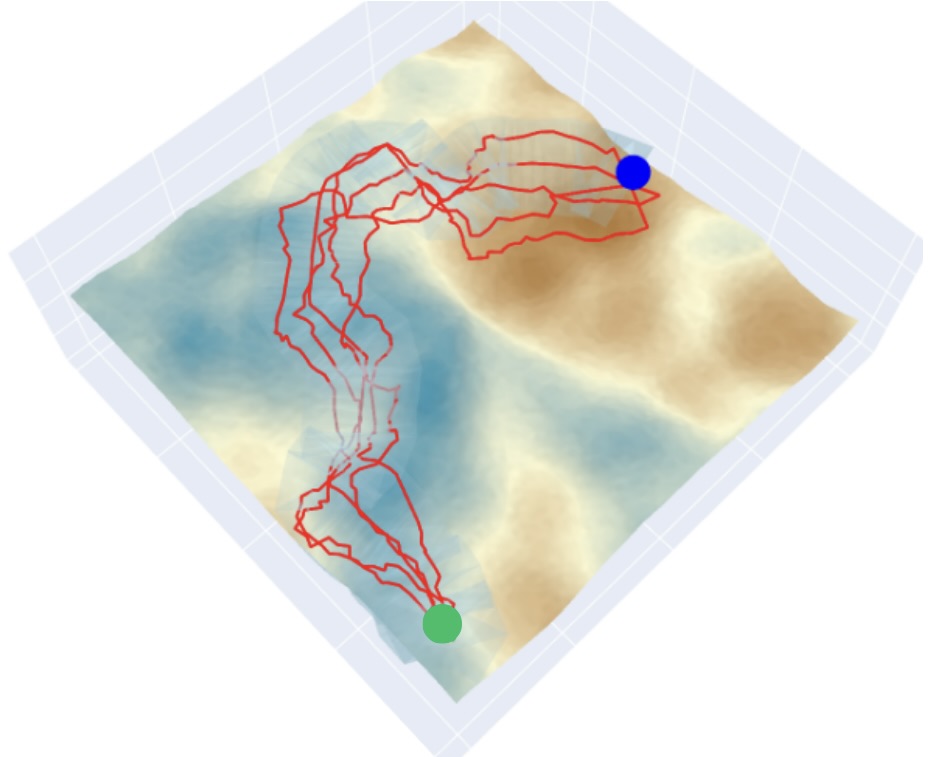}
        \caption{
         The trajectory of an Unmanned Surface Vehicle (USV) from Sec \ref{sim_expts} attempting to navigate from a \textcolor{mygreen}{starting point (green dot)} to a \textcolor{blue}{goal point (blue dot)}. The USV's \textcolor{red}{route (red lines)} visibly adjusts in response to the changing wave heights.}
        \label{fig:tv-vis}
    \end{subfigure}
    \caption{Time-varying environments and their effects.}
    \label{fig:combined}
\end{figure}

Consider an autonomous marine robot deployed to traverse an unfamiliar environment over an extended period of time.  
Such a task is often complicated by the environmental conditions that are subject to time-varying forces such as waves, tides
or thermohaline circulation 
\cite{WolfFlather2005, geo, ElHawary2000}. 
Given that the environment is subject to change, it is often reasonable to assume that these changes occur at a relatively slow pace as illustrated in Fig. \ref{fig:figure1}. 
These features can  generate significant uncertainty whose effects on the robot's dynamics are challenging to accurately predict or model \cite{long-d, jmse11061164}.
Consequently, the robot's dynamics can be considered unknown, time-varying and stochastic. 

While POMDPs provide a framework for sequential decision making under uncertainty, the introduction of time-variability in POMDP is nontrivial. 
As \cite{sukha} highlights, the challenge lies in the fact that time is unidirectional and incorporating time into state space leads to an explosion in the size of the state space, dramatically increasing computational complexity.  
To address this, we propose \textit{Time-Varying POMDPs} (TV-POMDPs) that represent transitions as dynamic probability functions avoiding state space explosion. 

The key implication is that the robot's dynamics at any given time depends on the current state and the temporal context. The effects of an action taken in a particular state may have different outcomes depending on when it is executed.
This makes it challenging to learn policies for sequential decision making and control, since past experience may become obsolete under time-variation.
As shown by \cite{sukha}, a basic Bellman backup mechanism \cite{bellman1957markovian} is not sufficient. This presents a twofold challenge: accurately estimating the time-varying dynamics of the system and effectively planning the system's actions in response to this dynamic uncertainty. 
We propose \textit{Memory Prioritized State Estimation} (MPSE), a strategy which hinges on the understanding that in an environment characterized by incremental change, historical data still holds considerable value to learn the transition function --- there exists a substantial correlation between past observations, recent observations and the current state.

\subsection{Related Works}
Classical approaches like MDPs \cite{Bert05} and POMDPs \cite{kaelbling:aij98} offer valuable foundations for modeling dynamics and decision-making under uncertainty. However, these models assume stationary transition probabilities, which are not suitable for time-varying environments. 

Time-Dependent MDPs (TDMDPs) \cite{NIPS2000_09b15d48} encode time  as a component in the state space.
However, analysis \cite{timd} shows that solving TDMDPs can be computationally intractable, limiting real-world use. 
Semi-MDPs \cite{SUTTON1999181} allow only transition times to vary while keeping probabilities stationary.
$\epsilon$-stationary MDPs \cite{KalmarSzepesvariLorincz1998, SzitaTakacsLorincz2002} only permit small variations in probabilities, insufficient for dynamic environments. 

An approach better suited to time-varying environments is
Time-Varying MDPs (TVMDPs) \cite{sukha}, that accommodates spatial and temporal stochasticity in transitions. 
However, they rely on MDPs that are known a priori. Further work \cite{mornik} extends this to learn and plan in unknown TVMDPs. A key limitation is that these do not incorporate partial observability, which is ubiquitous in real-world settings.

Learning in time-varying environments poses significant challenges. Stationary RL methods \cite{Watkins1992} \cite{NIPS1999_464d828b} require extensive training episodes to converge to optimal behavior. However, in time-varying settings, the environment continues to evolve during training data collection, invalidating previously learned policies \cite{icmlw, drift}. 
Online RL algorithms \cite{sutton2018reinforcement} adapt policies continuously, but naively updating estimates risks decreased performance if past experiences become obsolete under time-variation \cite{lange2012batch, pnas}.

Retaining pertinent memories is key for estimation and planning in time-varying environments \cite{RAMP200968}. Memory-based approaches \cite{memo, posts} can outperform Monte Carlo \cite{nipu} methods but resetting memory when dynamics change is unsuitable for time-variation.
Selective experience replay \cite{see} and memory attention \cite{memo2} allow replaying pertinent experiences.  However, these may still replay outdated experiences if temporal relevance is not considered.

Our work builds on these approaches by adaptively prioritizing observations based on temporal relevance, facilitating estimation and planning under time-varying, partial observability.

\subsection{Contributions }
In this paper, we introduce a Time-Varying Partially Observable Markov Decision Process (TV-POMDP) framework to model time-varying stochastic dynamics.
We propose Memory Prioritized State Estimation (MPSE) to facilitate estimation and planning within this framework. 
MPSE selectively prioritizes observations based on correlation and recency. 
We derive a modified Maximum Likelihood Estimate using these prioritized samples to optimize state estimation.
Additionally, we present a planning strategy tailored to leverage the identified time-varying transitions to improve action selection.
We demonstrate our approach in real-world and simulated experiments, considering an autonomous marine vehicle traversing waypoints and a robot navigating stochastic, time-varying environments.
 Our results show improved state estimation accuracy and superior planning performance, highlighting the effectiveness of our proposed method. 

\section{Preliminaries}
\label{sec:pre}
\subsection{Partially Observable Markov Decision Process}
A Partially Observable Markov Decision Process (POMDP) \cite{kaelbling:aij98} is an extension of the classic Markov Decision Process (MDP) that is used to model decision-making problems in which the agent cannot directly observe the underlying state of the system. 
It is represented by the tuple $(S, A, Z, T, O, R, \gamma)$, which denotes state space, action space, observation space, state transitions, observation functions, rewards, and the discount factor, respectively.

The agent uses a \textit{belief state}, which is the probability distribution over the state space \( S \) and is updated as the agent interacts with the environment and receives observations. 
The agent's aim within a POMDP is to determine an optimal policy \( \pi^*\), optimizing the expected cumulative rewards. 

\section{Time-Varying Partially Observable Markov Decision Process}
POMDPs model the environment as a stochastic process for optimal decision-making, assuming stationary transition probabilities. However, real-world dynamics like those of the autonomous marine robot that was introduced earlier are often time-varying due to varying factors like winds and waves.
Fig. \ref{fig:tv-vis} provides an illustration of a  scenario where the robot's transition function is subject to temporal variations. 
To better model such scenarios, we propose an extension to the traditional POMDP framework.

Our modified POMDP retains the original state and action spaces, avoiding inflation in computational complexity. 
The TV-POMDP model is represented by the tuple $(S, A, Z, T_t, O, R_t, \gamma, b_0)$ capturing state, action, and observation spaces, time-varying transition and reward functions, the observation function, the discount factor, and the initial belief state $b_0$ at time  \(t_0\).  
The \textit{time-varying transition probability}, denoted by $T_t(s, a, s')$, defines the probability of transitioning from state $s$ to state $s'$ under action $a$ at time $t$.

Similarly, the reward function is reformulated as $R_t(s,a)$ which gives the immediate reward for taking action $a$ in state $s$ at time $t$. 
The value function $V$, traditionally representing the 
cumulative discounted return
from belief state $b$, now incorporates the temporal dimension --- $V_t(b)$ gives the expected future discounted reward from belief state $b$ at time $t$. 
The value function $V_t(b)$ is given by:
\begin{equation}
V_t(b) = \max_{a \in A}\sum_{s \in S} b(s) \left[ R_t(s, a) + \gamma \sum_{s' \in S} T_t(s', s, a) V_{t+1}(b') \right],
\label{vtb}
\end{equation}
where $\gamma$ is the discount factor and $V_{t+1}(b')$ is the value of belief state $b'$ at future time $t+1$.

\section{Learning and Planning in a TV-POMDP}
The TV-POMDP formulation captures the temporal dynamics of the environment and acknowledges that the same action taken in the same state can lead to different outcomes at different points in time, i.e., \textit{non-repeatability} of outcomes. 
For instance, the tides might be stronger at
certain times of the day compared to others --- negating the possibility of the robot coming back to the same state and learning from multiple runs.
This feature necessitates the development of an online learning approach. 

However, online learning in the described context has its own set of intricacies. 
If the transition probabilities were independent, i.e., the observations at time $t-1$ do not impact the estimate of transition probabilities for time $t$, it is meaningless to attempt to learn the transition probability function.
Thus learning needs to leverage some structural characteristics.
In many practical domains, there exists some innate knowledge regarding the maximal rate of environmental changes, $\Delta_{max}$,
\begin{equation*}
|T_t(s',a|s) - T_{t-1}(s',a|s)| \leq \Delta_{\text{max}}.
\end{equation*}
We leverage this underlying structure to serve as a foundation for the agent's learning strategy, guiding its process of estimating and planning in such time-varying environments. 
Our problem, thus, is a two-fold one: 
\begin{enumerate}
    \item Online learning: Estimate $T_t$ in a single run 
    by interacting with the environment, leveraging the knowledge of its maximal rate of change.
    \item Optimal decision making: Using $T_t$, develop an optimal policy $\pi^*$ that maximizes the future rewards.
\end{enumerate}

\subsection{Memory Prioritized State Estimation}
In order to learn and estimate a transition probability function, we need to rely on information in form of observations from the environment. 
Processing all past observations is computationally expensive, so we focus on \textit{windows} of recent observations 
with the premise that, compared to individual observations, these windows can provide a comprehensive view of the time-varying dynamics at play. The size of the time window is empirically chosen, and the method is generally not very sensitive to the window size as long as it is within a range that captures the temporal patterns.

\subsubsection{Memory Prioritization} 
Not all observations  carry equal significance. Some might offer better insights into the temporal patterns and transitions in the environment, while others might be relatively uninformative, imperfect or incomplete. 
We propose a prioritizing scheme that assigns a weight to each sample based on its potential \textit{informativeness}. 
This weight is determined by a combination of factors:
\begin{enumerate}
\item \textit{Autocorrelation}: Quantifies the degree to which observations are dependent on each other, measuring serial dependence. 
For a series of observations $z = (z_1, z_2, \ldots, z_n)$, the autocorrelation \cite{autoc} at lag $k$ is:
\begin{equation}
\rho(k) = \frac{\sum_{t=k+1}^{n} (z_t - \bar{z})(z_{t-k} - \bar{z})}{\sqrt{\sum_{t=1}^{n} (z_t - \bar{z})^2 \sum_{t=1}^{n} (z_{t-k} - \bar{z})^2}},
\end{equation}
We define the autocorrelation score $A_s$ by 
\begin{equation}
A_s(z_i) = \rho(t - t_i). 
\end{equation}
\( A_s(z_i) \) captures how similar the dynamics of the environment are at time \( t \) to those at time \( t_i \) when \( z_i \) was observed. High value of \( A_s(z_i) \) suggest that \( z_i \) is pertinent and can provide insight for decision-making at \( t \).

\item \textit{Recency}: The value of an observation often diminishes over time, captured by a recency score \( R_s \). For an observation \( z_i \) at time \( t_i \), $R_s$ is formulated as
\begin{equation}
R_s(z_i) = \frac{1}{t - t_i + \varepsilon},
\end{equation}
where \( t \) is the current time and \( \varepsilon \) prevents division by zero. \( R_s \) quantifies the freshness of an observation. Higher \( R_s \) means the observation is more recent and more likely to be relevant for the current state of the system.

\item \textit{Deviation}: Observations that differ significantly from the mean, signifying critical transitions. 
\begin{equation}
D_s(i) = |z_i - \bar{z}_t|.
\end{equation}
Here, \( D_s \) quantifies the extent to which an observation \( z_i \) deviates from the mean \( \bar{z}_t \) within its time window. While higher $D_s$ values can indicate both informative observations and outliers, the inclusion of autocorrelation score reduces the potential impact of outliers, allowing $D_s$ to be less influential on the overall estimation.
\end{enumerate}

Each observation $z_i$ is assigned a combined weight $\omega_i$, which is a weighted sum of these factors:
\begin{equation}
\omega_{z_i} =  w_a{A_s(z_i)} + w_r{R_s(z_i)} + w_d{D_s(z_i)},
\end{equation}
where $w_a$, $w_r$ and $w_d$ are the weights that represent the relative importance of autocorrelation, recency and deviation.

\subsubsection{Estimation of Time-Varying Transition Probability Function}
To estimate the time-varying transition probability function, we use a weighted likelihood approach informed by our prioritized memory scheme.
We define a likelihood function \cite{mornik}, \( L(T_t|o_{0:\mathcal{T}}, a_{0:\mathcal{T}}, \omega_{0:\mathcal{T}}) \), that encapsulates the probability of observing a particular sequence of observations \( z = (z_0, z_1, \ldots, z_\mathcal{T}) \) within a window, given a sequence of actions \( a = (a_0, a_1, \ldots, a_\mathcal{T}) \) and the model \( T_t \).
\begin{equation}
L(T_t|z_{0:\mathcal{T}}, a_{0:\mathcal{T}}, \omega_{0:\mathcal{T}}) = \prod_{t=0}^{\mathcal{T}} \left[ \mathrm{Prob}_{\hat{P}}(z_t|a_t, \omega_t) \right]^{\omega_t},
\end{equation}
\( \mathrm{Prob}_{\hat{P}}(z_t|a_t, \omega_t) \) represents the probability of observing \( z_t \) given action \( a_t \) and the model, adjusted by the weight \( \omega_t \).

To solve for \( T_t \), we frame it as a constrained optimization problem with the objective of maximizing the log-likelihood function, subject to maximal rate of change constraint \( \Delta_{\text{max}} \).

The optimization problem is then:
\[
\begin{aligned}
\hat{T}_t(o,a,o') &= \arg\max_{T_t} \text{ }\log L(T_t|o_{0:\mathcal{T}}, a_{0:\mathcal{T}}, \omega_{0:\mathcal{T}}) \\
&\text{subject to} \quad |T_t(o',a|o) - T_{t-1}(o',a|o)| \leq \Delta_{\text{max}}.
\end{aligned}
\]
The work in \cite{mornik} shows that this is a convex problem.
To solve this problem efficiently in real-time, we employ CVXOPT \cite{Andersen2012}, an optimization library.

\subsubsection{Policy Optimization and Planning}
Our ultimate objective is to devise an optimal policy using the estimated transition function, $\hat{T}_t(s,a,s')$, that maximizes expected long-term rewards. To reflect the system's dynamic nature, we adjust the Bayesian belief state update with the estimated $T_t$:
\begin{equation}
b_{t+1}(s') = \eta \Omega(z_t|s_t',a_t) \sum_{s \in S} \hat{T}_t(s_t',a_t|s_t)b_t(s).
\end{equation}
Here, $b_{t+1}(s')$ is the belief state at time $t+1$, $\Omega(z|s',a)$ is the likelihood of observing $z$ after taking action $a$ and transitioning to state $s'$, and $\eta$ is a normalizing constant to ensure a valid probability distribution. 
Optimal policy, $\pi(b)$, leverages expected $V_t(b)$ (Equation \ref{vtb}) to maximize expected rewards:
\begin{equation}
\pi(b) = \arg\max_{a \in A} \sum_{s' \in S} R_t(s,a) \hat{T}_t(s,a,s') b(s) + \gamma V_t(b').
\label{final_eq}
\end{equation}
The problem's complexity in Equation \ref{final_eq} is tied to the dimensions of the state and action spaces, the real-time optimized transition function \( \hat{T}_t(s,a,s') \) significantly reduces the computational burden. Commonly used algorithms such as Policy Iteration is adept at solving this problem efficiently.

\begin{figure*}[!htb]
    \centering
    \begin{subfigure}[b]{0.23\textwidth}
        \centering
        \includegraphics[width=\textwidth, trim={30 30 20 10},clip]{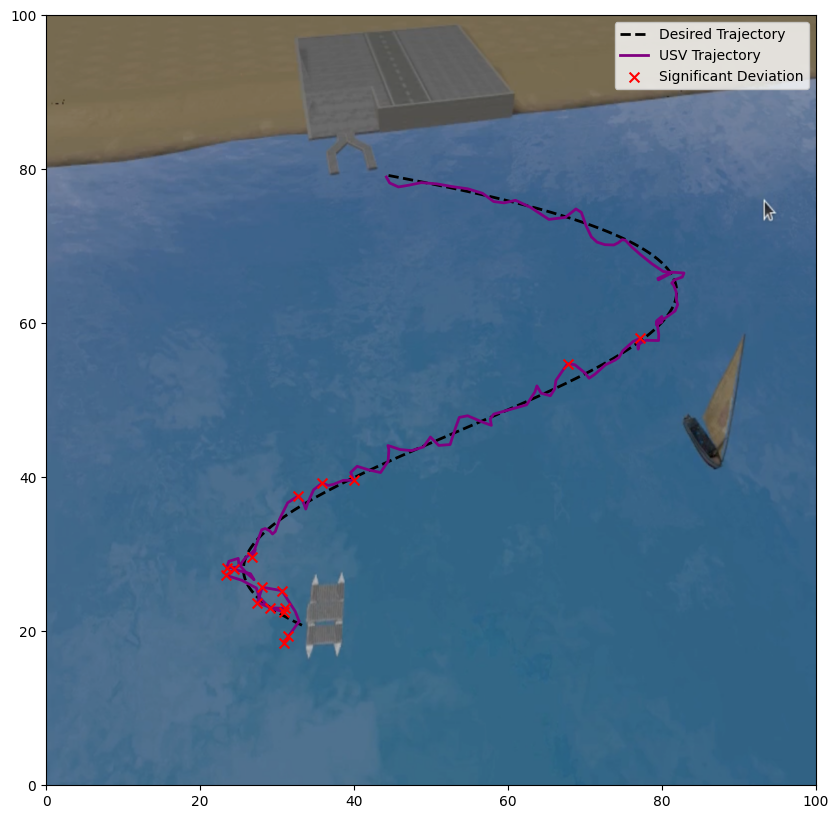}
        \caption{Our approach (MPSE): \\ \# waypoints  followed: 117}
    \end{subfigure}
    \hfill
    \begin{subfigure}[b]{0.23\textwidth}
        \centering
        \includegraphics[width=\textwidth, trim={30 30 20 10},clip]{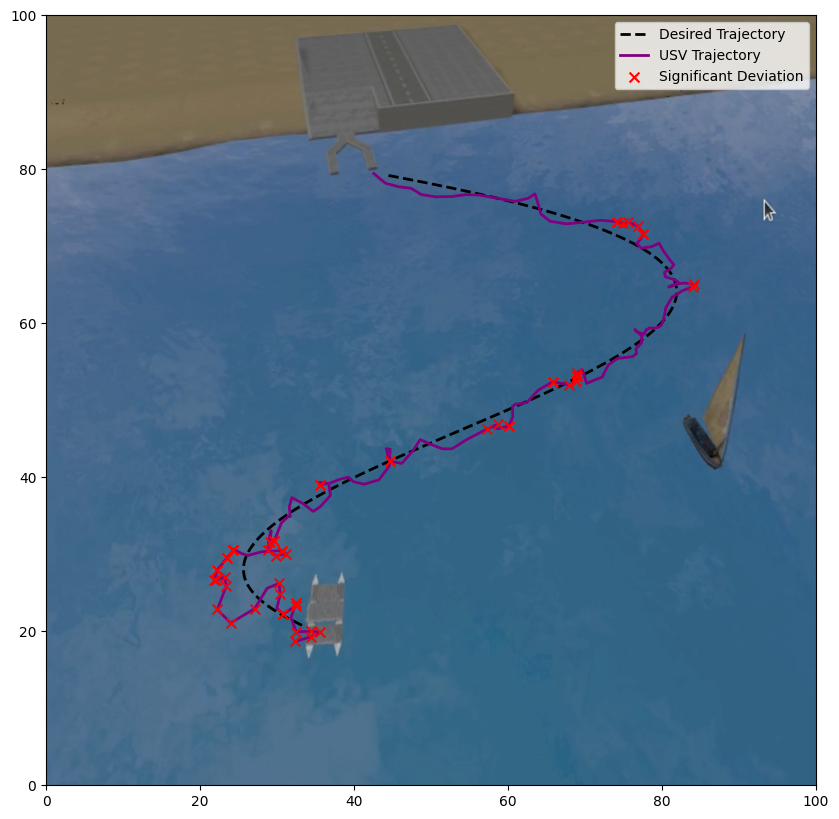}
        \caption{DT-POMDP (classical estimation): \# waypoints followed: 103}
    \end{subfigure}
    \hfill
    \begin{subfigure}[b]{0.23\textwidth}
        \centering
        \includegraphics[width=\textwidth, trim={30 30 20 10},clip]{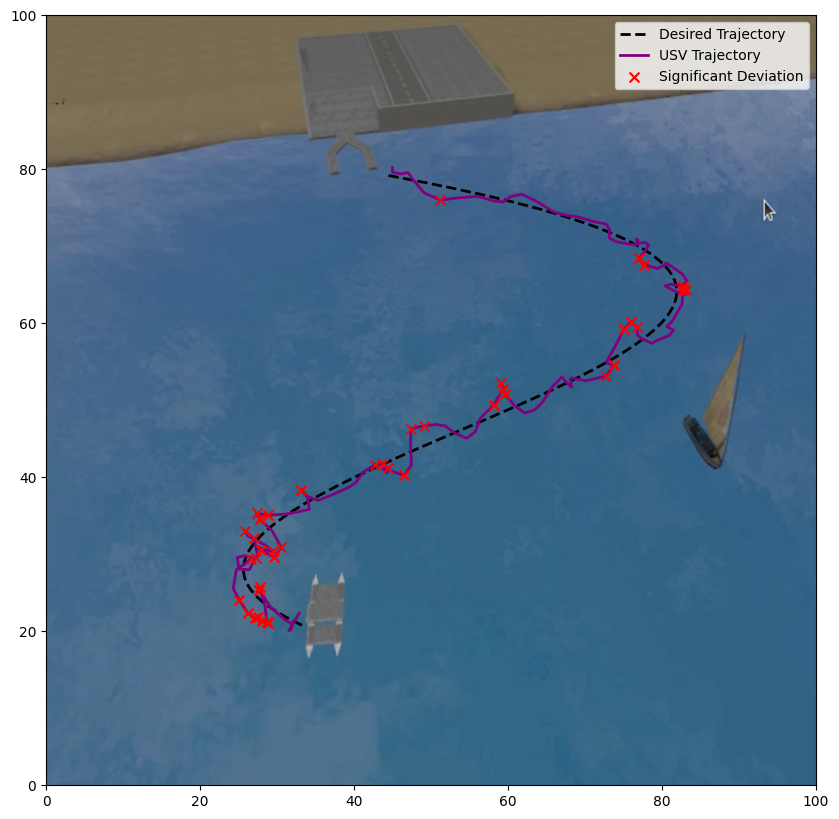}
        \caption{TA-DESPOT: \\ \# waypoints followed: 99}
    \end{subfigure}
    \hfill
    \begin{subfigure}[b]{0.23\textwidth}
        \centering
        \includegraphics[width=\textwidth, trim={30 30 20 10},clip]{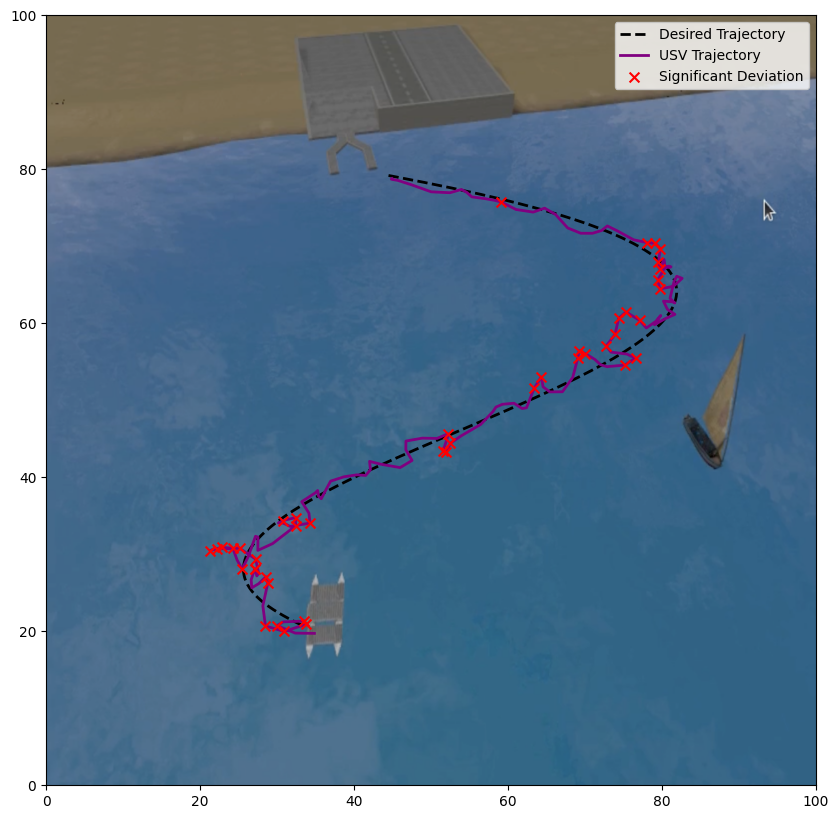}
        \caption{TA-PBVI: \\ \# waypoints followed: 99}
    \end{subfigure}
    \caption{
        The figures display the simulation environment and the waypoints tracked by the algorithms within a TV-POMDP (Scenario~2 with \(\Delta_{\text{max}}=0.02\)).  If the robot's estimate from a waypoint exceeds 3m, it is marked with \textcolor{red}{\textbf{x}} indicating notable deviations. The \textcolor{darkpurple}{purple lines} indicate the USV's trajectory, and the \textcolor{black}{black lines} are the desired 150-waypoint trajectory.
    }
    \label{marine}
\end{figure*}

\section{Experiments and Discussions}
This section evaluates our algorithm's performance in TV-POMDPs through two cases: a simulated marine environment with an USV following waypoints, and a robot in a hardware experiment navigating the shortest path to a destination.
\subsection{Simulated Marine Experiment} \label{sim_expts}
We simulate an USV navigating 150 waypoints, with the ultimate goal of reaching a final waypoint. The simulation is carried out in Gazebo \cite{gazebo} with ROS2 \cite{ros2} as the middleware. Adopted from \cite{bingham19toward}, the simulation models sophisticated water-vessel interactions through hydrodynamics plugin and utilizes the wavefield plugin \cite{tessy} to reproduce effects of winds and waves. 
\begin{itemize}
    \item \textbf{Action Space:} The USV, based on \cite{BORISOV2016256}, 
is modeled as a single engine vessel with action space \(\mathcal{A}\) defined by:
    \begin{itemize}
        \item Thrust $\in[0, 35]N$, with discrete increments of 5N
        \item Thrust Direction $\in [-90^\circ,90^\circ]$, with \( 10^\circ \) increments.
    \end{itemize}
    
    \item \textbf{Partial Observations:} The USV receives noisy position estimates as observations, with noise distributed by $\sim \mathcal{N}(0, \sigma^2)$
where \(\sigma=1.5 \text{m}\) was adjusted to reflect USV GPS accuracy levels \cite{gpsusv}.

\item \textbf{Reward:} \( \textstyle 
\begin{cases}
    +10, & \parbox[t]{.6\linewidth}{\text{per waypoint reached in the right}\\ \text{order within a radius of \( 3 \)m}} \\
    +50, & \text{upon reaching the final goal}
    \\ 0, & \text{otherwise}
    
\end{cases} \)

    \item \textbf{Model Uncertainty:} 
    \label{modlu}
    Hydrodynamic disturbances introduce time-varying transitions in simulation. Specifically, if a disturbance occurs, the trajectory deviates by an angle  $\theta \sim \mathcal{N}(0, 10^2), \, \theta \in [-10^\circ, 10^\circ]$.
\end{itemize}

\subsubsection{Baselines}
The standard POMDP model does not natively handle time-varying dynamics, limiting applicability. We introduce modifications to enable comparisons.

We use classical estimation techniques with the Discrete-Time POMDP (DT-POMDP), which adds discrete time elements to create time-layered state structures. However, DT-POMDP has fixed horizons and discretization limitations.
We extend Point-Based Value Iteration (PBVI) \cite{pbvi}, a point-based POMDP planning method, to Time-Aware PBVI (TA-PBVI) by incorporating a time dimension into the belief points.
We also extend Determinized Sparse Partially Observable Tree (DESPOT) \cite{despot}, a belief tree planning method, to Time-Aware DESPOT (TA-DESPOT) by incorporating time steps as state variables during tree expansion to handle time.

\subsubsection{Simulation results}
We evaluate the performance of the algorithms in two distinct scenarios:

\underline{Scenario~1 - Constant Environmental Conditions}:
We employ the wavefield plugin to simulate a constant wind speed of \(5 \, \text{m/s}\), thereby fixing the transition probability at $0.7$ --- essentially operating in a standard POMDP.  For any action, the USV transitions to the expected next state with 70\% probability. With 30\% probability, a disturbance occurs, deviating the trajectory as described in \ref{modlu}.
This simplified scenario provides a performance reference before introducing time-varying disturbances in Scenario~2.

\begin{figure*}[!htb]
    \centering
    \begin{subfigure}[b]{0.23\textwidth}
        \centering
        \includegraphics[width=\textwidth]{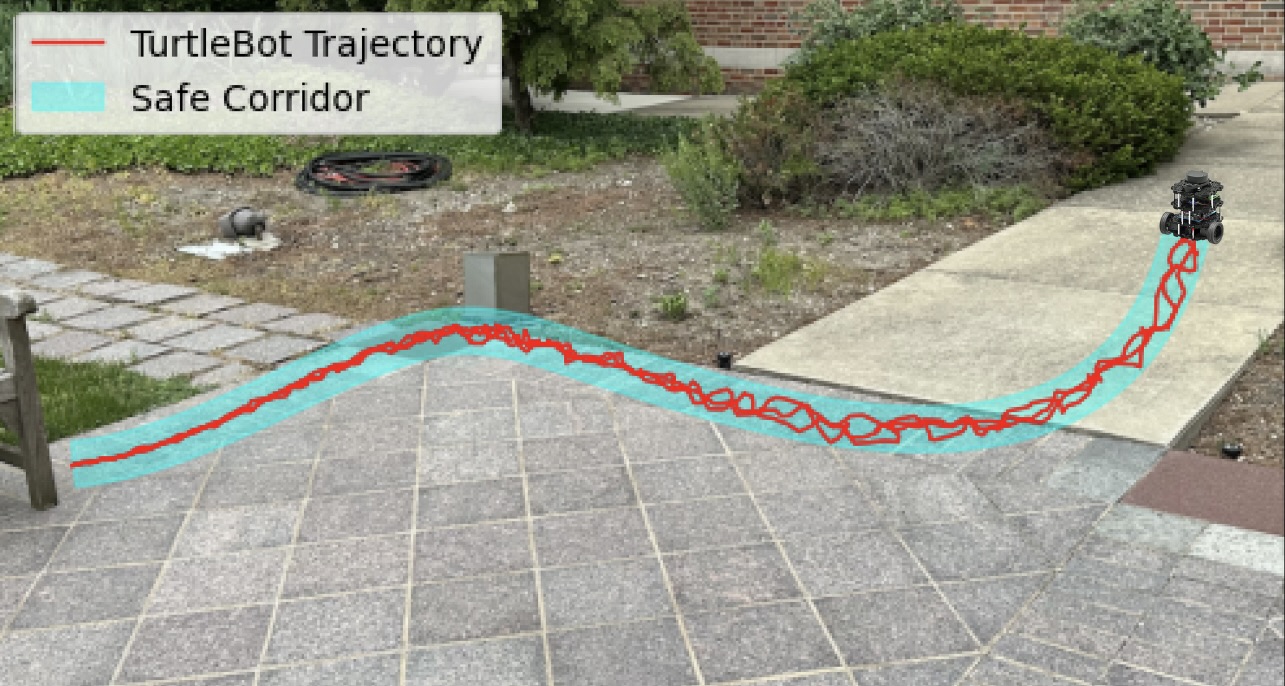}
        \caption{MPSE}
    \end{subfigure}
    \hfill
    \begin{subfigure}[b]{0.23\textwidth}
        \centering
        \includegraphics[width=\textwidth]{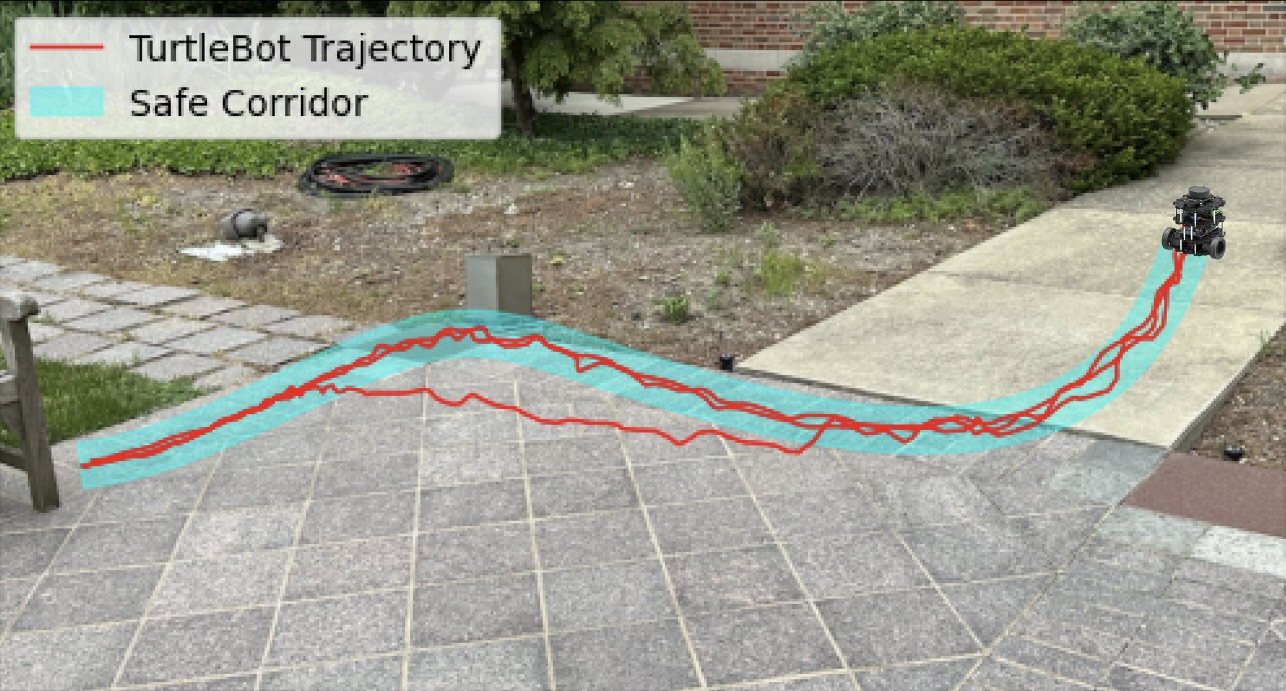}
        \caption{DT-POMDP}
    \end{subfigure}
    \hfill
    \begin{subfigure}[b]{0.23\textwidth}
        \centering
        \includegraphics[width=\textwidth]{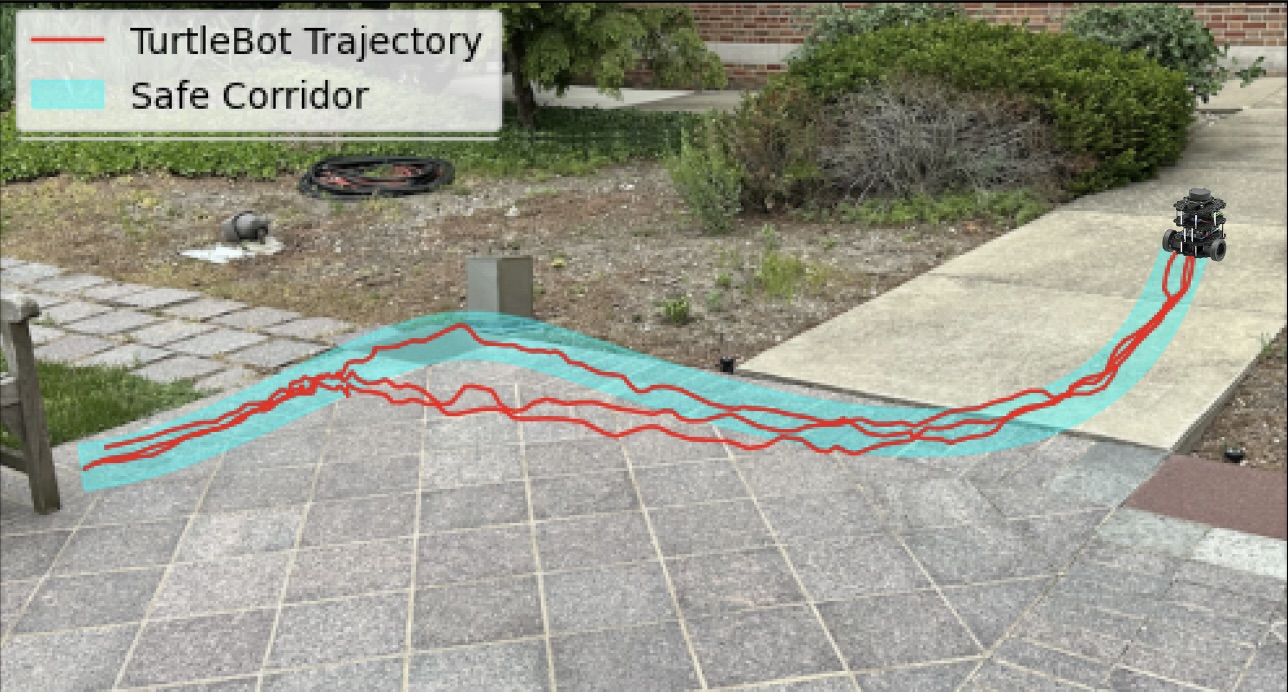}
        \caption{TA-DESPOT}
    \end{subfigure}
    \hfill
    \begin{subfigure}[b]{0.23\textwidth}
        \centering
        \includegraphics[width=\textwidth]{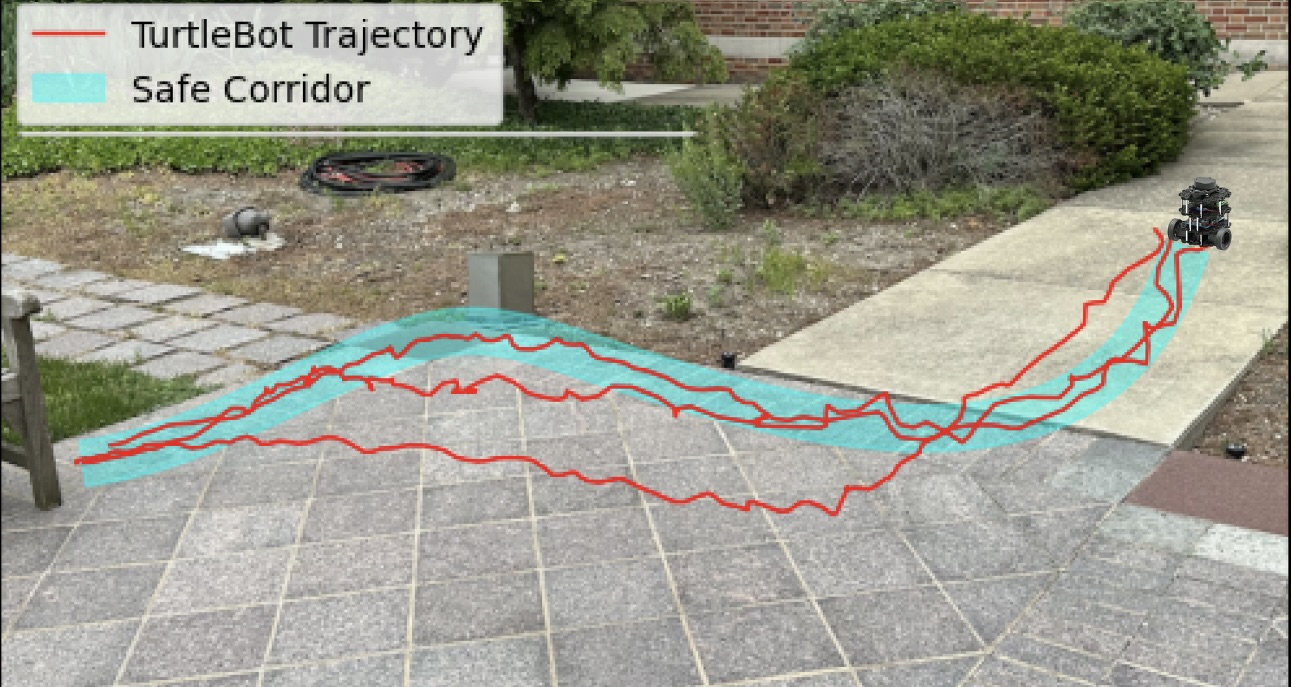}
        \caption{TA-PBVI}
    \end{subfigure}
    \caption{
        The figures represent the hardware setup and trajectories for MPSE and baselines in a TV-POMDP (Scenario~3 with \(\Delta_{\text{max}}=0.03\)). \textcolor{cyan}{Cyan corridor} represents the map limits (\(H_{\text{safe}}\)) and \textcolor{red}{red lines} represent TurtleBot's trajectories. The background consists of two different terrains: the white areas depict rougher cemented terrain, while the darker areas show smoother terrain. The roughness variation between these surfaces is substantial enough to impact the TurtleBot's mobility.
    }
    \label{turtle}
\end{figure*}

\underline{Scenario~2 - Adding time-varying disturbances}:
We use the hydrodynamics plugin and the wavefield  plugin to introduce a time-varying transition function,
$T_t = 1 - t/20$. 
This function simulates increasing currents over time that hinder the USV's motion, causing more stochastic transitions \cite{geo}. The USV's trajectory deviates by a probability of $t/20$ as described in \ref{modlu}. The linear current growth matches real-world acceleration of currents in certain climates \cite{ana}. 

\begin{table}[h]
    \centering
    \renewcommand{\arraystretch}{1.2} 
    \begin{tabular}{c V{2.5} c c V{2.5} c c}
    \arrayrulecolor{black}\hlineB{2.5}
    \rowcolor{Gray}
    & \multicolumn{2}{c|}{\begin{tabular}[c]{@{}c@{}}\textbf{Constant Transition} \\ \textbf{Probability}\\ \(T_t = 0.7\) \\ for \(0 \le t \le 20\)\end{tabular}} & \multicolumn{2}{c}{\begin{tabular}[c]{@{}c@{}}\textbf{Time-Varying} \\ \textbf{Transition Probability} \\ \(T_{t} = 1-t/20\) \\ for \(0 \leq t \leq 20\) \\ \(\Delta_{\text{max}} = 0.02 \)\end{tabular}} \\
    \arrayrulecolor{black}\clineB{2-5}{2.5}
    \rowcolor{LightGray}
    & \begin{tabular}[c]{@{}c@{}}\textbf{MAE}\end{tabular} & \begin{tabular}[c]{@{}c@{}}\textbf{\# waypoints} \\ \textbf{followed} \end{tabular} & \begin{tabular}[c]{@{}c@{}}\textbf{MAE}\end{tabular} & \begin{tabular}[c]{@{}c@{}}\textbf{\# waypoints} \\ \textbf{followed} \end{tabular} \\
    \arrayrulecolor{black}\hlineB{2.5}
    \textit{\textbf{MPSE}} & 0.16626 & 121 & \textbf{0.22820} & \textbf{117} \\
    \hline
    DT-POMDP & 0.19505 & 110 & 0.25921 & 103 \\
    \hline
    TA-DESPOT & \textbf{0.14221} & \textbf{129} & 0.27003 & 99 \\
    \hline
    TA-PBVI & 0.16088 & 123 & 0.29612 & 99 \\
    \arrayrulecolor{black}\hlineB{2.5}
    \end{tabular}
    \vspace{5mm} 
    \caption{Comparison of mean absolute error (MAE) and the number of waypoints followed for Scenarios 1 and 2. The USV transitions to the next expected state with probability $T_t$ and deviates with probability $1-T_t$ as described in \ref{modlu}.}
    \label{tab:comparison_sim}
\end{table}

\underline{Discussions:}
As shown in the results of Scenario~1 (Table \ref{tab:comparison_sim}, the baselines are well-established for solving standard POMDPs. 
TA-DESPOT performs best due to its exhaustive look-ahead search, while MPSE closely mirrors the performance. 
As shown in our simulations (Fig. \ref{marine}) and the results (Table \ref{tab:comparison_sim}), the time-varying dynamics in Scenario~2 present significant challenges for established baselines despite using time as a variable in their models. 
Unlike our approach, the baselines do not include mechanisms to prioritize historical samples alongside new data. This leads to their core algorithms' inability to adapt effectively to evolving, uncertain dynamics, highlighting a critical limitation: a sole reliance on the most recent observations is not adequate for accurate planning in time-varying POMDPs. On the contrary, MPSE's selective combination of past and present information enables more effective adaptation to unknown, evolving dynamics. 

Without mechanisms to prioritize samples from memory alongside new data, it becomes difficult for their core algorithms to adapt to the unknown, evolving dynamics. 
Fundamentally, this underscores a key limitation of 
relying solely on the most recent observations to accurately plan in a time-varying POMDPs. 
MPSE's superior performance highlights the value of this mechanism for settings with unknown, continuously changing dynamics. 

\subsection{Hardware Ground Vehicle Experiment}

To further evaluate the capability of our algorithm in real-world settings, we conduct physical experiments using a TurtleBot Burger navigating a rough terrain.
The robot is tasked with navigating from a designated start point to a predefined goal location in the shortest possible route while ensuring that it does not go off-road. 
The TurtleBot is equipped with an onboard computer, u-blox ZED F9P GPS module, and LIDAR for SLAM \cite{slam}. The agent has access to the safe coridor ($H_{\text{safe}}$) through an occupancy grid ($H$), generated using SLAM. $H_{\text{safe}}$ defines the shortest, safest corridor to the objective. 
It moves at a maximum speed of 0.22 m/s, receiving 10 Hz GPS observations. We leverage ROS to interface the sensors with the onboard computer.
\begin{figure}[ht]
\centering
\includegraphics[trim=0 0 0 0, clip, width=0.35\textwidth]{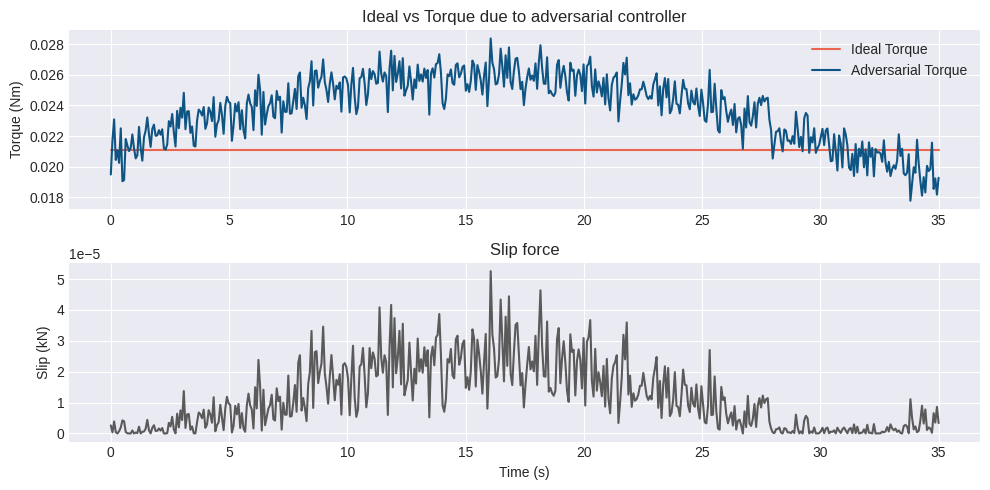}
\caption{The \textcolor{red}{red line} represents the robot's maximum torque, while the \textcolor{blue}{blue line} illustrates the effect of the introduced adversarial controller. The \textcolor{darkbrown}{brown line} depicts the resulting slippage on the robot's wheels. To empirically validate these effects on the model's transition probabilities, we conducted trials with and without the adversarial controller. By tracking the robot's GPS transitions and comparing against the ground truth without adversarial control, we tuned the model to account for the torque and slippage impact on transitions.}

\label{fig:torque}
\end{figure}

\begin{itemize}
\item \textbf{Action Space:} The robot's movements are defined by its steering angle, captured in the discretized action space:
\begin{itemize}
\item Steering Angle ($\theta$): $\theta \in [-\varphi,\varphi]$ in increments of $5^\circ$, where $\varphi \in [-90,90]$ is the maximum steering angle.

\end{itemize}

    \item \textbf{Partial Observations:} 
    At each timestep, noisy GNSS provides the position estimate modeled as
        signal strength $SS_1 = f(\hat{d}_t)$, where $\hat{d}_t$ is the true position plus Gaussian noise $\epsilon \sim \mathcal{N}(0, \sigma^2)$ with horizontal error $\sigma=1.5m$ \cite{gnss}.

\item \textbf{Reward Function:} The reward \( R(h) \) at any point \( h \) in the occupancy grid \( H \) is defined as follows:
\[
R(h) = 
\begin{cases} 
-1, & \text{if } h \in H_{\text{safe}}, \\
-10, & \text{if } h \in H_{\text{unsafe}},\\
+5, & \text{on reaching the objective waypoint}
\end{cases}
\]

\item\textbf{Model Uncertainty:}
\label{modlu2}
To emulate time-varying environmental conditions like increasing rain, which could cause slip, we implement an adversarial controller that modulates the wheel torque --- simulating the varying levels of wheel-ground friction, mimicking the effect of slip on the state transitions of the robot. Fig. \ref{fig:torque} shows the effects of this artificially induced, time-varying slip. Consequently, the robot's trajectory deviates within an angle range \( \zeta \sim \mathcal{N}(0, \sigma^2) \) where \( \zeta \in [-10^\circ, 10^\circ] \) and $\sigma = 5^\circ$.

\end{itemize}

\begin{table*}[!htb]
    \centering
    \renewcommand{\arraystretch}{1.2} 
    \begin{tabular}{c c c | c c | c c}
    \arrayrulecolor{black}\hlineB{2.5}
    \rowcolor{Gray}
    & \multicolumn{2}{c|}{\begin{tabular}[c]{@{}c@{}}\textbf{Constant Slip}\\ $T_t = 0.9$ for $0 \le t \le 20$\end{tabular}} & \multicolumn{2}{c|}{\begin{tabular}[c]{@{}c@{}}\textbf{Time-Varying Slip}\\ $T_{t}= 0.9 \times e^{-0.05t}$ \\ $\text{ for } 0 \leq t \leq 20 \text{ and } \Delta_{max} = 0.02 $\end{tabular}} & \multicolumn{2}{c}{\begin{tabular}[c]{@{}c@{}}\textbf{Time-Varying slip}\\ $T_{t}=  \begin{cases} 
    1/(1 + e^{- (t - 5)}); 0\le t \le 10\\
    e^{-(t - 10)^2/{2}}; 10\le t \le 20 \end{cases}$\end{tabular}} \\
    \arrayrulecolor{black}\clineB{2-7}{2.5}
    \rowcolor{LightGray}
    & \textbf{MAE} & \textbf{Cost incurred} & \textbf{MAE} & \textbf{Cost incurred} & \textbf{MAE} & \textbf{Cost incurred} \\
    \arrayrulecolor{black}\hlineB{2.5}
    \textit{\textbf{MPSE}} & 0.09652 & -34 & \textbf{0.14201} & \textbf{-45} & \textbf{0.18195} & \textbf{-61}  \\
    \hline
    DT-POMDP & 0.11505 & -56 &  0.18569 & -62 & 0.30395 & -84  \\
    \hline
    TA-DESPOT &\textbf{ 0.07221} & \textbf{-31} & 0.23723 & -81 & 0.39528 & -86  \\
    \hline
    TA-PBVI & 0.09988 & -43 & 0.27453 & -96 & 0.41741 & -102  \\
    \arrayrulecolor{black}\hlineB{2.5}
    \end{tabular}
    \vspace{5mm} 
    \caption{Comparison of total error and total cost incurred for different scenarios. The TurtleBot transitions to next expected state with the defined probability $T_t$ and deviates with a probability $1-T_t$ as described in \ref{modlu2}.}
    \label{tab:comparison}
\end{table*}

\begin{figure*}[!htb]
    \centering
\includegraphics[width=0.8\linewidth, trim={0 0 0 28pt}, clip]{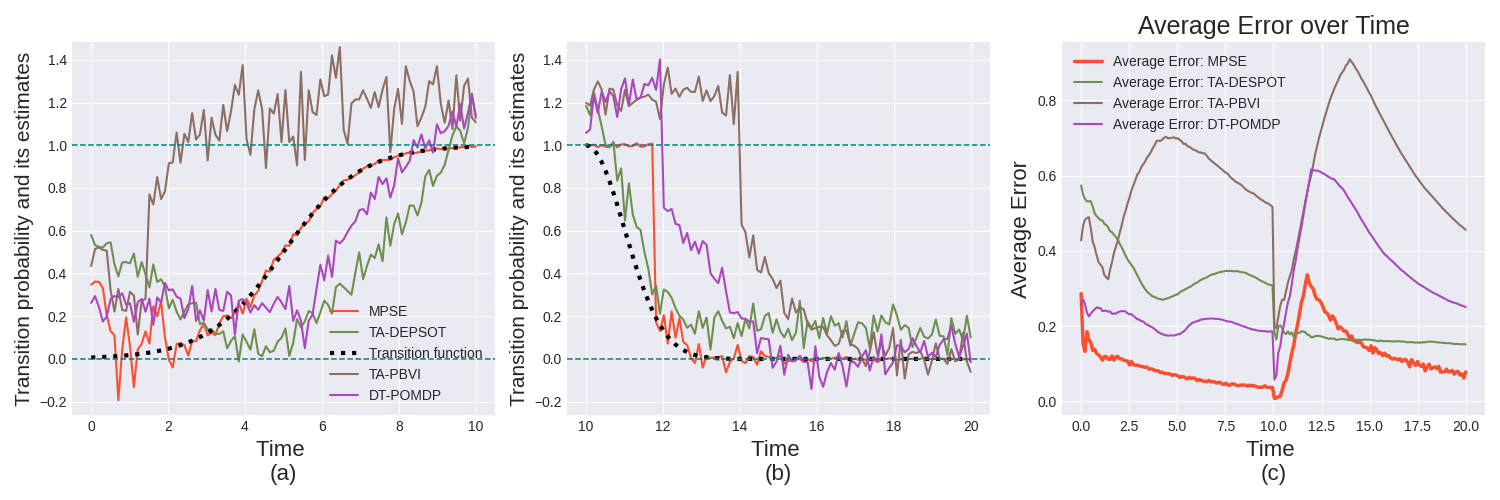}
    \caption{Subfigures (a) and (b) display transition function changes in the first and second mission halves, respectively, while (c) shows maximal error over time. To ensure valid probabilities, the estimated transition probabilities are clipped between \textcolor{teal}{teal lines}. The results depicted are after applying this clipping constraint. Although clipping alters individual probability values, it does not significantly impact the overall trends and comparative results. The key conclusions remain valid.}
    \label{fig:trans_est}
\end{figure*}

\subsubsection{Hardware results}
The results presented in Table \ref{tab:comparison} quantify the total error and the cost incurred when navigating from the start to goal in three different scenarios. Scenario setting for 1 and 2 remain consistent with the simulation settings. 
As shown, the results for the first two scenarios roughly align with the simulation results discussed earlier. 

\underline{Scenario~3: Adaptive Transition Functions:}
We challenge our approach's adaptability by switching the transition functions midway through the experiment --- emulating abrupt change in terrain. The robot transitions to the next expected state following \( T_{t1}(t) = {1}/({1 + e^{- (t - 5)})} \) for the first 10s, after which it switches to \( T_{t2}(t) = e^{-{(t - 10)^2}/{2}} \) for the next 10s. Fig. \ref{turtle} presents the trajectories followed in this scenario and Fig. \ref{fig:trans_est} presents the estimated transition function.

\underline{Discussions:}
The third scenario significantly challenges all the algorithms.
DT-POMDP with classical estimation struggled due to inability to swiftly adjust, a limitation in non-stationary settings \cite{opap}. 
TA-PBVI and TA-DESPOT were constrained by their dependence on preset beliefs/trees that are difficult to update. This intrinsic design choice means that when the environment changes unexpectedly, they have to heavily recompute or adjust these structures. This process is computationally intensive and time-consuming, leading to less agile responses.
As observed in Fig. \ref{fig:trans_est}, TA-DESPOT responded fastest owing to its flexible tree \cite{despot}. However, its convergence lagged as optimizing the tree under altered conditions is difficult \cite{nipu}.
MPSE took time to balance retaining prior learning versus acquiring new samples, a tradeoff in lifelong learning \cite{PARISI201954}. But after this adjustment period, MPSE  outperformed all methods in convergence speed through continuous adaptation. 
By continuously balancing learning with selective forgetting, MPSE provides robust performance despite time-varying transitions.

\section{Conclusion}
This work introduced Time-Varying Partially Observable Markov Decision Processes (TV-POMDPs) to address the challenges of decision-making in time-varying environments. To facilitate effective planning in such environments, we proposed a novel methodology called Memory Prioritized State Estimation (MPSE). Motivated by the need to manage the uncertainty arising from time-varying dynamics and partial observability, MPSE selectively prioritizes observations based on their information content, thereby enhancing the accuracy of estimations and ensuring computationally efficient planning.
An associated planning strategy was also presented, tailored specifically to exploit the modeled time-varying transitions. We validated our approach through real-world and simulated experiments demonstrating the effectiveness of modeling the relevant systems as TV-POMDPs and utilizing MPSE for more accurate decision-making. Our planning strategy outperformed multiple baselines in leveraging the time-varying nature of the environment for optimized action selection.
Our proposed work successfully integrates capabilities to plan in stochastic, partially observabile, and time-varying environments, overcoming limitations inherent in previous approaches.

\bibliographystyle{IEEEtran}
\bibliography{references}

\end{document}